\documentclass[10pt,journal,compsoc]{IEEEtran}
\usepackage{graphicx}
\usepackage{ragged2e}
\usepackage{todonotes}
\usepackage{hyperref}
\usepackage{cleveref}
\usepackage{multirow}
\usepackage{paralist}
\usepackage{listings}
\usepackage{mathtools}
\usepackage{amsfonts}
\usepackage{amssymb}
\usepackage{amsmath}
\usepackage{footnote}

\usepackage{ftnxtra}
\usepackage{tabularx}
\usepackage{multirow}
\usepackage{rotating}
\usepackage{paralist}
\usepackage{fnpos} 
\usepackage{cite}

\usepackage{rotating}
\usepackage{subcaption}
\usepackage{todonotes}
\usepackage{footmisc}
\usepackage{hyperref}
\usepackage{booktabs}
\usepackage{cleveref}

\begin{document}
\title{Convolutional Embedded Networks for Population Scale Clustering and\\ Bio-ancestry Inferencing}

\author{Md. Rezaul Karim$^{^{\star 1,2}}$, Michael Cochez$^{^3}$, Achille Zappa$^{^4}$, Ratnesh Sahay$^{^4}$, Oya Beyan$^{1,2}$,\\ Dietrich Rebholz-Schuhmann$^{^5}$, Stefan Decker$^{1,2}$

\IEEEcompsocitemizethanks{\IEEEcompsocthanksitem $^1$Fraunhofer Institute for Applied Information Technology FIT, Germany\protect\\
                \IEEEcompsocthanksitem $^2$RWTH Aachen University, Aachen,  Germany\protect\\
                \IEEEcompsocthanksitem $^3 $Dept. of Computer Science, Vrije Universiteit Amsterdam, Netherlands\protect\\
                \IEEEcompsocthanksitem $^4$Insight Centre for Data Analytics, NUI Galway, Ireland\protect\\
                \IEEEcompsocthanksitem $^5$German National Library of Medicine, University of Cologne, Germany
                }}


\markboth{IEEE/ACM Transactions on Computational Biology and Bioinformatics,~Vol.~XXX, No.~XXX, March~20XX}%
{Karim \MakeLowercase{\textit{et al.}}: Convolutional Embedded Networks for Population Scale Clustering and Bio-ancestry Inferencing}
\IEEEtitleabstractindextext{%
\begin{abstract} \justifying The study of genetic variants~(GVs) can help find correlating population groups to identify cohorts that are predisposed to common diseases and explain differences in disease susceptibility and how patients react to drugs. Machine learning algorithms are increasingly being applied to identify interacting GVs to understand their complex phenotypic traits. Since the performance of a learning algorithm not only depends on the size and nature of the data but also on the quality of underlying representation, deep neural networks~(DNNs) can learn non-linear mappings that allow transforming GVs data into more clustering and classification friendly representations than manual feature selection. In this paper\footnotetext{Under review in IEEE/ACM Transactions on Computational Biology and Bioinformatics}, we proposed convolutional embedded networks~(CEN) in which we combine two DNN architectures called convolutional embedded clustering~(CEC) and convolutional autoencoder~(CAE) classifier for clustering individuals and predicting geographic ethnicity based on GVs, respectively. We employed CAE-based representation learning on 95 million GVs from the `1000 genomes'~(covering 2,504 individuals from 26 ethnic origins) and `Simons genome diversity'~(covering 279 individuals from 130 ethnic origins) projects. Quantitative and qualitative analyses with a focus on accuracy and scalability show that our approach outperforms state-of-the-art approaches such as VariantSpark and ADMIXTURE. In particular, CEC can cluster targeted population groups in 22 hours with an adjusted rand index~(ARI) of 0.915, the normalized mutual information~(NMI) of 0.92, and the clustering accuracy~(ACC) of 89\%. Contrarily, the CAE classifier can predict the geographic ethnicity of unknown samples with an F1 and Mathews correlation coefficient~(MCC) score of 0.9004 and 0.8245, respectively. To provide interpretations of the predictions, we identify significant biomarkers using gradient boosted trees~(GBT) and SHapley Additive exPlanations~(SHAP). Overall, our approach is transparent and faster than the baseline methods, and scalable for 5\% to 100\% of the full human genome.
\end{abstract}

}

\maketitle
\IEEEdisplaynontitleabstractindextext
\IEEEpeerreviewmaketitle

\section{Introduction}
\label{section1}
Genetic variations~(GVs) are structural variation in the DNA sequence in human genomes, which makes us all unique in terms of phenotype, e.g., genetic polymorphism is implicated in numerous diseases and constitute the majority of varying nucleotides among human genomes~\cite{6}. Genetic research has played a significant role in the discovery of new biological pathways underpinning complex human disease and the evaluation of new targets for therapeutic development~\cite{bomba2017impact}. In particular, biological understanding of the relationship between GVs can provide insights into the disease status of a patient~(e.g., interacting GVs contributing to breast cancer risk)~\cite{behravan2018machine}, which is the prerequisite to enable personalized treatment. Further benefits of studying GVs lies in the discovery and description of the genetic contribution to many human diseases based on their haplogroup and ethnicity~\cite{byun2017ancestry}. Subsequently, finding similar population groups and identifying patients who are predisposed to common or rare diseases at early stages is increasingly important in understanding the effects of biomarkers on the development of certain disease~\cite{miller2001understanding}. 

Despite strides in characterizing human history from GVs data, progress in identifying genetic signatures of recent demography has been limited. Predicting haplogroup and ethnicity accurately~(called bio-ancestry inferencing~(BAI)\footnote{Geographic origin of an individual based on genetic variants}) is very challenging in which one of the most critical tasks is the analysis of genomic profiles to attribute individuals to specific ethnic populations or the interpretation of nucleotide haplotypes for diseases susceptibility \cite{laitman2013haplotype}. Consequently, BAI is a frequently studied problem, with the main goal of identifying an individual’s population of origin based on our knowledge of natural populations~\cite{padhukasahasram2014inferring}. Accordingly, BAI has numerous applications in forensic analyses, genetic association studies, and personal genomics. Further, BAI is used as a checksum method to verify a sample’s integrity, e.g., case-control studies~\cite{byun2017ancestry}, where BAI is key to understand population stratification across the cohorts to help avoid probable spurious associations with even subtle ancestry differences~\cite{sheehan2016deep}. Identification of ancestry is an important research challenge in which any direct assessment of disease-related GVs will yield more insights~\cite{31}. 

Since the race of an individual depends on ancestry, grouping each into a cluster is expected to correlate with the traditional concepts of race. However, this correlation is not perfect since GVs occurs according to probabilistic principles, which often does not follow any continuous distribution across races but slightly overlaps across diverse populations, e.g., research~\cite{30} has exposed that population groups from Asia, Europe, Africa, and America can be separated based on their genomic data based on the fact that `Y' chromosome lineage can be geographically localized, forming an evidence for clustering of human alleles. In this study, we try to understand: i) if individual genetic profiles can be used to attribute into specific populations, ii) if individual GVs are suitable for predicting it's ethnicity, iii) how individuals are distributed geographically, e.g, among population groups. However, grouping individuals or BAI based on GVs requires access to  genomics data and efficient analytic methodologies to cope with millions of GVs from thousands of individuals~\cite{2,3,4,5}. However, extracting significant features from a large-scale GVs is not only computationally expensive but also a critical bottleneck. 

Previous approaches~\cite{ADMIXTURE,21} employed machine learning~(ML) based approaches to answer these questions. In particular, ML-based clustering algorithms such as DBSCAN, OPTICS, Gaussian Mixture~(GMM), Agglomerative clustering~(AC), and K-means have long been used in literature to address issues with higher-dimensional input spaces. However, they are fundamentally limited to linear embedding~\cite{28}. Further, since structure of GVs data is different than numeric or categorical data, non-linear embedding is necessary before initializing the clustering operation. Hence, ML-based approaches failed to exploit non-linear relationships from high-dimensional GVs data and exhibit good accuracy at BAI and clustering tasks. In contrast, approaches based on neural networks~(DNNs) can be more effective at RL and feature extraction \cite{karim2020BiB}. In particular, a DNN architecture~(e.g., autoencoder~(AE)) with multiple hidden layers and non-linear activation functions, can  capture more complex and higher-level features and contextual information from the input~\cite{jaques2017multimodal}. Further, non-linear mappings allows transforming input data into more clustering-friendly representations in which the data is mapped into a lower-dimensional feature space that helps fine-tune clustering~\cite{min2018survey}. 

Although DNN have shown tremendous success to deal with such complex tasks, they are mostly perceived as `black box' methods because of lack of understanding of their functionalities~\cite{karim2019onconetexplainer}, which is a serious drawback. To recommend more accurate treatments and drug repositioning, interpretability is essential to provide insights on why and how a certain prediction is made by the algorithm outlining important biomarkers. Further, the `right to explanation' of EU GDPR~\cite{kaminski2019right} gives patients the right to know why and how an algorithm made a diagnosis decision. Hence, DL-based systems have to be GDPR complaint. 

We try to address the challenges and requirements in a scalable and efficient ways: first, we use Spark and ADAM for processing large-scale GVs to convert them into genotype objects. Then convolutional autoencoder~(CAE) is employed  representation learning on GVs data. Learned features are then used to: i) train the convolutional embedding clustering~(CEC) for clustering individual to determine inter and intra-population groups, ii) train the CAE classifier to predict ethnicity of unknown samples. Finally, to provide interpretations of the predictions, we identify significant biomarkers using gradient boosted trees~(GBT) and SHapley Additive exPlanations~(SHAP). The rest of the paper is structured as follows: \cref{re} discusses related works and analyze their potential limitations. \Cref{section3} chronicles our proposed approach in detail with materials and methods. \Cref{section4} demonstrates some experiment results, discuss the findings, and highlights potential limitations of the study. \Cref{section6} provides some explanations of the importance and relevance of the research reported and discussed some future works before concluding the paper.

\section{Related Work} 
\label{re}
Although 1000GP consortium has developed a global reference for human genetic variation for exome and genome sequencing and despite strides in characterizing human history from genetic polymorphism data, progress in identifying genetic signatures of recent demography has been limited~\cite{han2017clustering}. Lek M. et al.~\cite{27}, describe aggregation and analysis of high-quality protein-coding region and DNA sequence data for 60,706 individuals of diverse ancestries in which the objective metrics of pathogenicity are calculated for sequence variants against various classes of mutations. Their approach: i) identify as much as 3,230 genes with near-complete depletion of predicted protein-truncating variants, while 72\% of these genes have no currently established human disease phenotype, ii) demonstrate that GVs can be used for efficient filtering of candidate disease-causing variants, which helps in the discovery of human `knockout' variants in protein-coding genes.

Studies on population structure clustering include ~\cite{han2017clustering}, which identify fine-scale population structure clustering of 770,000 genomes in North America, which reveals post-colonial population structure. Another used approach is ADMIXTURE~\cite{ADMIXTURE}, which performs maximum likelihood estimation~(MLE) of individual ancestries from multilocus SNP genotype data. However, this approach cannot cluster GVs comfortably giving an ARI of only 0.25. Further, ADMIXTURE requires a preprocessing step from VCF to PED format, which takes a significant amount of time. To address the shortcomings of ADMIXTURE, VariantSpark is proposed~\cite{21}, which provides an interface from MLlib that offers a seamless genome-wide sampling of variants and provides a pipeline for visualizing results from the 1000GP and PGP. However, overall clustering accuracy is low, and VariantSpark does not provide support for classifying individuals based on genotypic information. 

Research also focused on genomic inferring, and ethnicity prediction, e.g., literature ~\cite{pritchard1999population} proposed approximate Bayesian computation~(ABC), which is a likelihood-free inference method based on simulating datasets and comparing their summary statistics. Although ABC's main advantages lie in its simplicity and ability to output a posterior distribution, it suffers from the `curse of dimensionality' with decreasing accuracy and stability as the number of summary statistics grows~\cite{sheehan2016deep}.
Jinyoung B. et al~\cite{byun2017ancestry}, proposed a distance-based approach for BAI using principal component analysis and spatial analysis to assign individuals to population memberships. 

In a previous work~\cite{karimdeep}, we applied K-means for the population scale clustering and achieved better accuracy than ADMIXTURE and VariantSpark. For predicting ethnicity, we trained an MLP classifier, which achieved a state-of-the-art result with high confidence. However, two limitations remained: i) the feature extraction process based on SPARQL query and converting genotype data into Resource Description Format (RDF)~\cite{1} take non-trivial time for all the chromosomes. Excellent performance was obtained for the genotype dataset for a single chromosome due to a low number of latent variables, which shows inferior results for all the chromosomes because of the overfitting and lack of generalization while training MLP model.

\section{Materials and Methods}
\label{section3}
In this section, we describe materials and methods of our approach: first, we describe our feature engineering step we followed to prepare the training data consisting of GVs features and labels. Then we chronicle network constructions for clustering and classifying population groups. Finally, we describe the training process and hyper-parameter tuning. 

\subsection{Problem statement}
Clustering individual's based on GVs is correlated with geographic ethnicity and bio-ancestry, where the main objective is grouping populations into clusters based on similarity, density, intervals, or particular statistical distributions measures of the data space~\cite{karim2020BiB}. Given GVs of $n$ samples, $X$ = ${\mathbf{\{x_1,x_2, ..., x_n}}\}$, where $X \in \mathbb{R}^{D}$. We consider clustering individuals into $k$-categories~(i.e. $k$ super-population or sub-population groups), each represented by a centroid $\mu_{j}, j=1, \dots, k$. On the other hand, predicting the ethnicity of an individual $x_i$ is classifying a data point into a specific into sub-population groups based on its GVs.

However, instead of clustering or classifying samples directly in the original data space $X$, we first transform the data with a nonlinear mapping $f_{\theta}: X \rightarrow Z$, where $\theta$ are learnable parameters and $Z \in \mathbb{R}^{K}$ is the learned or embedded feature space, where $K \ll D$. In our approach, to parametrize $f_{\theta}$, CAE architecture is employed due to it's function approximation properties and feature learning capabilities~\cite{xie2016unsupervised} by capturing  local relationship values~\cite{KarimDDI2019} in GVs with convolutional~(conv) filters.  

\begin{figure*}
	\centering
	\includegraphics[width=\textwidth]{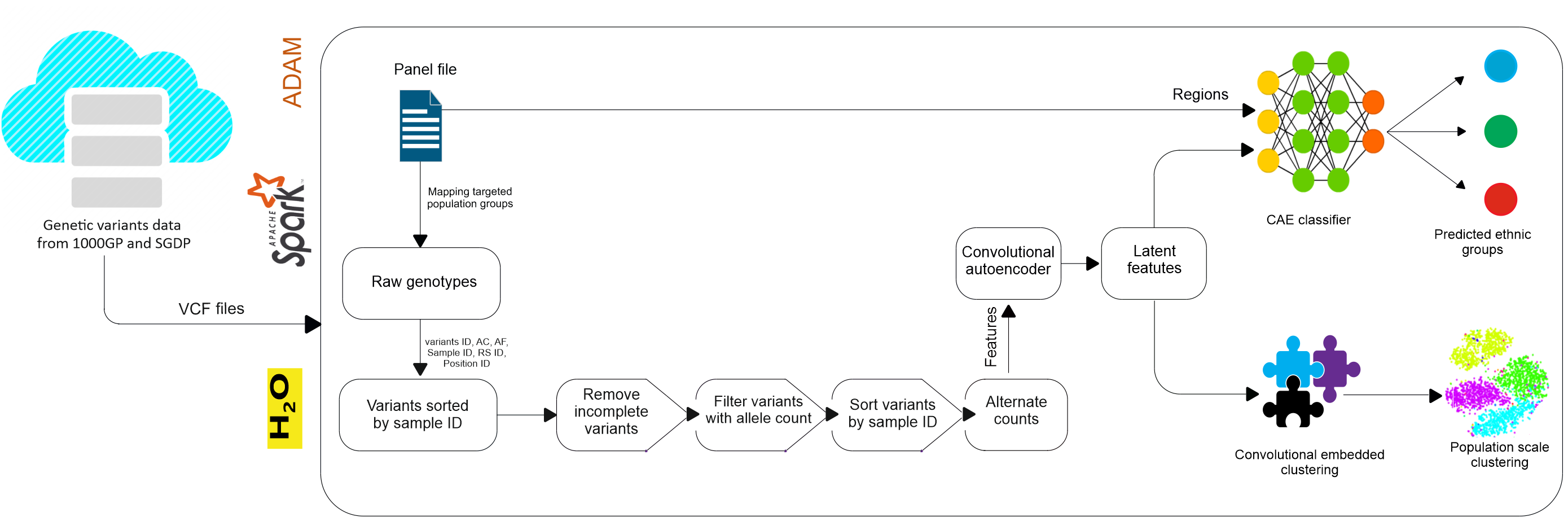}
	\caption{processing pipeline of the proposed approach for for population-scale clustering and bio-ancestry inferencing}	
	\label{fig:wf}
\end{figure*}

\subsection{Datasets}
\label{section2}
Various genomics projects based on next Generation Sequencing technologies have emerged, including The Cancer Genome Atlas~(TCGA)\footnote{\textbf{TCGA}: \url{https://cancergenome.nih.gov/}}, International Cancer Genome Consortium~(ICGC)\footnote{\textbf{ICGC}: \url{https://dcc.icgc.org/}}, 1000 genomes project\footnote{\textbf{1000GP}: \url{https://www.internationalgenome.org/}}~(1000GP), Human Genome Project~(HGP)\footnote{\textbf{HGP}: \url{http://humangenes.org/}}, Simons Genome Diversity Project~(SGDP)\footnote{\textbf{SGDP}: \url{http://reichdata.hms.harvard.edu/sgdp/}}, and Personal Genome Project(PGP)\footnote{\textbf{PGP}: \url{http://www.personalgenomes.org/}}.
The HGP showed that significant genetic differences exist between individuals, whereas, inspired by HGP, 1000GP seeks to measure those differences to help biomedical researchers understand the roles of GVs in health and illness. 
In previous studies, the 1000GP~\cite{6, 10002015global} serves as one of the prime sources to analyze genome-wide single nucleotide polymorphisms~(SNPs) at scale for predicting individual's ancestry with regards to continental and regional origins~\cite{belsare2019evaluating}.

\subsubsection{Data selection}
Data used in this study from the 1000GP~(phase 3) and SGDP act large catalog of human GVs, where the phase 3 of 1000GP provides GVs data of 2,504 individuals from 26 populations~(i.e., ethnicity) in which samples are grouped into five super-population groups according to their predominant ancestry: Europe, Africa, America, and Asia in which each of the 26 populations has about 60-100 individuals~\cite{39}.

\begin{table*}
\centering
	\caption{distribution of 1000GP population groups}
	\label{SPG}
	\vspace{-3mm} 
	\begin{tabular}{l|l|l|l}
		\hline
		 \textbf{Super population group} & \textbf{Acronym} & \textbf{Covering ethnicity} & \textbf{Sample count}\\ \hline
		East Asia & EAS & CHB, JPT, CHS, CDX, KHV & 617  \\
        European & EUR & CEU, TSI, FIN, GBR, IBS & 669 \\
        African & AFR & YRI, LWK, GWD, MSL, ESN, ASW, ACB & 1,018  \\
        American & AMR & MXL, PUR, CLM, PEL & 535 \\
        South Asian & SAS & GIH, PJL, BEB, STU, ITU & 661 \\
		\hline
	\end{tabular}
\end{table*}

However, deletions and substitutions of less important variants~(single nucleotide polymorphisms~(SNPs), indels, and other structural variant classes) in quality control leaves a total of 88 million variants: 84.7 million were SNPs, 3.6 million short indels, and 60,000 structural variants) identified as high-quality haplotypes~\cite{6, 10002015global, karim2018scala}. Each genotype comprises all 23 chromosomes and a separate panel file containing samples and population information. For multi-allelic variants~(e.g., Listing:1), each alternative allele frequency~(AF) is calculated as the quotient of allele count and allele number~(AN), where AF in the five super-population groups is calculated from the AN(range=[0,1]).

\begin{table*}
    \centering
	\caption{a snapshot from the panel file from 1000GP}
	\vspace{-3mm} 
	\begin{tabular}{l|l|l|l|l}
		\hline
		\textbf{Sample ID} & \textbf{Population group} & \textbf{Ethnicity} &	\textbf{Super population group} & \textbf{Gender} \\
		\hline
		HG00096	& GBR & British in England and Scotland & EUR & Male \\
		HG00171	& FIN & Finnish in Finland	& EUR & Female \\
		HG00472	& CHS & Southern Han Chinese & EAS & Male \\
		HG00551	& PUR & Puerto Ricans from Puerto Rico & AMR & Female \\
		\hline
	\end{tabular}
	\label{table3}
\end{table*}

However, one downside of 1000GP is that it's sequencing study focused on demographically large populations, which, unfortunately, tend to ignore smaller populations that are also important for understanding human diversity~\cite{chakravarti2015perspectives}. To include such smaller populations, GVs from the SGDP are used, which contains deep genome sequences of 279 individuals from 130 populations chosen to span much of human genetic, linguistic, and cultural variation, covering: 44 Africans, 22 Native Americans, 27 Central Asians or Siberians, 47 East Asians, 25 Oceanians, 39 South Asians, and 75 West Eurasians. The 1000GP\footnote{\url{ftp://ftp.1000genomes.ebi.ac.uk/vol1/ftp/release/20130502/}} and SGDP data\footnote{\url{https://reichdata.hms.harvard.edu/pub/datasets/sgdp/}} are publicly available in variant call format~(VCF). Additionally, population region for each sample is provided.

\vspace{3mm}
\begin{minipage}[c]{\linewidth}
        \begin{lstlisting}[captionpos=b, caption={an example of multi allelic variants in 1000GP}, label=mutli_allelic_var_data, basicstyle=\ttfamily,frame=single] 
        1 15211 rs78601809 T G 100 PASS AC=3050;AF=0.609026;AN=5008;NS=2504;DP=32245; EAS_AF=0.504;AMR_AF=0.6772;AFR_AF=0.5371;EUR_AF=0.7316;SAS_AF=0.6401; AA=t|||;VT=SNP
    \end{lstlisting}
\end{minipage}

\subsubsection{Population stratification} 
Since majority of the variants are SNPs and INDELs~\cite{sheehan2016deep}, computing likelihood of complex population genetic models is often infeasible from the multiple individuals. Further, population stratification is necessary to identify the presence of a systematic difference in AF between sub-populations in a population, possibly due to different ancestry. First, we process the panel file containing sample IDs of the individuals, population group, ethnicity, super population groups, and the gender info~(\cref{table3}) and extract only the targeted population data which identify the population groups. Then we convert the GVs to common genotype object, followed by another round of filtering to extract data for the relevant individuals and super population groups only. Genotype objects are then converted into a sample variant object containing the following genotypic information: 

\vspace{1mm} 
\begin{itemize}
    \item \textbf{Sample ID:} to uniquely identify a sample
    \item \textbf{Variant ID:} to uniquely identify a genetic variant
    \item \textbf{Alternate allele count:} count of alternate alleles~(AA) in which the sample differs from the reference genome. 
\end{itemize}
\vspace{1mm} 

Furthermore, since ADMIXTURE’s underlying statistical model does not take linkage disequilibrium~(LD) into account, we remove variants with high LD and incomplete variants, assuming they are outliers~\cite{variantspark}. Moreover, since 1000GP phase 3 contains overlapping and duplicate sites, we ignored duplicate sites in any analysis. The total number of sample~(i.e., individual) is then determined, before grouping them using their variant IDs, and filtering out variants without support by the samples. Then we group variants by sample ID and sort them for each sample consistently using variant IDs, which gives us a sparse training data consisting of sample ID, variant ID, position ID, RS ID, and AA count, where a row represents an individual, a column represents a specific variant, and the ``Region'' column signifies labels. 

\subsection{Network constructions and training}
In our approach, we model two tasks into a single pipeline, which has three stages as shown in \cref{fig:wf}: i) representation learning~(RL) based on CAE, ii) population scale clustering using CEC in which we jointly optimize \emph{clustering} and \emph{non-clustering} losses, iii) ethnicity prediction using a CAE classifier. Non-clustering loss~(NCL) is independent of the clustering algorithm and usually enforces a desired constraint on the learned model~\cite{aljalbout2018clustering}. NCL also guarantees the learned representation to preserve spatial relationships between GVs so the original input can be reconstructed in the decoding phase~\cite{min2018survey}, while the clustering loss is specific to the clustering method and clustering-friendliness of the learned representations~\cite{aljalbout2018clustering}. 

\subsubsection{Construction of convolutional autoencoder}
\label{subsec:CAE_construct}
Inspired by literature~\cite{karim2020BiB}, we learn the cluster and classification friendly representations of samples by employing CAE: first, to apply conv operations, we embed extracted GVs of each sample into a 2D image inspired by literature \cite{KarimNCCA2019} in which each sample is reshaped from a 4,238x1 array into a 66 x 66 image by adding zero padding around the edges and normalized each image to [0,255]. Instead of manually engineered conv filters in a CNN, we constructed the CAE by adding conv and pooling layers. The CAE consists of an encoder that performs convolution and pooling operations and a decoder that performs unpooling and deconvolution~(deconv) operations. From the given GVs, the first conv layer in the encoder calculates $j^{th}$ feature map~(FM) as follows~\cite{alirezaie2019semantic}: 

\begin{equation}
    h^{j}=\sigma\left(x_{i} * W_{ij}^{j}+b^{j}\right),
\end{equation}

\noindent where $x_i$ is the input sample, $W_{ij}^{j}$ is the $j^{th}$ filter from input channel $i$ and filter $j$, $b^j$ is the bias for the $j^{th}$ filter, i.e., single bias per latent map\footnote{One bias per GV would introduce many degrees of freedom}, $\sigma$ is rectified linear unit~(ReLu) activation function, and $*$ denotes the conv operation. To obtain the translation-invariant representations, max-pooling is performed by downsampling conv layer's output and taking the maximum value in each $m \times n$ non-overlapping sub-region~\cite{alirezaie2019semantic}. 
In the decoding phase, unpooling and deconv operations are performed to preserve the positional-invariance information during the pooling operations. The deconv operation is performed to reconstruct $x_i$ as follows~\cite{alirezaie2019semantic}:

\begin{equation}
   x_i = \sigma\left(o^{j} * W_{oj}^{j}+c^{j}\right),
\end{equation}

\noindent where $o^j$ is $j^{th}$ FM and $W_{oj}^{j}$ is $j^{th}$ filter of unpooling layer $o$; $j$ and $c^j$ are filter and bias of $j^{th}$ output layer. Hence, CAE learns optimal filters by minimizing the RL1, which is the distance measure $d_{CAE}$ between input $x_i$ and its corresponding reconstruction $f(x_i)$: 

\begin{equation}
    L_{CAE}=\text{$d_{CAE}$}(x_i, f(x_i) = \sum_{i} ||x_{i}-f(x_i)||^{2}.
    \label{eq:Loss1}
\end{equation}

Eventually, CAE-based RL results more abstract features that help to stabilize training and network converges faster, avoid corruption in feature space, and improve clustering quality~\cite{karim2020BiB}. The architecture of CAE consist of a 20-layer network in which batch normalization layer is used after every conv layer and the ReLu activation is used in every layer except for the last layer where softmax activation is used. In particular, the CAE part has the following structure: 

\vspace{2mm} 
\begin{itemize}
    \item \textbf{Input layer}: genetic variants of each sample reduced from 4,238 $\times$ 1 to 66 $\times$ 66 $\times$ 1
    \item \textbf{Convolutional layer}: of size 32 $\times$ 32 $\times$ 32 
    \item \textbf{Batch normalization layer}: of size 32 $\times$ 32 $\times$ 32
    \item \textbf{Convolutional layer}: of size 16 $\times$ 16 $\times$ 64
    \item \textbf{Batch normalization layer}: of size 16 $\times$ 16 $\times$ 64
    \item \textbf{Max-pooling layer}: of size 2 $\times$ 2
    \item \textbf{Convolutional layer}: of size 8 $\times$ 8 $\times$ 128
    \item \textbf{Batch normalization layer}: of size 8 $\times$ 8 $\times$ 128
    \item \textbf{Convolutional layer}: of size 4 $\times$ 4 $\times$ 256
    \item \textbf{Batch normalization layer}: of size 4 $\times$ 4 $\times$ 256
    \item \textbf{Max-pooling layer}: of size 2 $\times$ 2
    \item \textbf{Upsampling layer}: of size 2 $\times$ 2
    \item \textbf{Deconvolutional layer}: of size 4 $\times$ 4 $\times$ 256
    \item \textbf{Batch normalization layer}: of size 4 $\times$ 4 $\times$ 256
    \item \textbf{Deconvolutional layer}: of size 8 $\times$ 8 $\times$ 128
    \item \textbf{Batch normalization layer}: of size 8 $\times$ 8 $\times$ 128
    \item \textbf{Deconvolutional layer}: of size 16 $\times$ 16 $\times$ 64
    \item \textbf{Batch normalization layer}: of size 16 $\times$ 16 $\times$ 64
    \item \textbf{Upsampling layer}: of size 2 $\times$ 2
    \item \textbf{Deconvolutional layer}: of size 32 $\times$ 32 $\times$ 32. 
\end{itemize}
\vspace{2mm}

\subsubsection{Construction and training of CEC network}
Architecturally, CEC is an improved variant of DEC in which we employed CAE instead of vanilla AE and denoising AE. The CEC is trained in two phases: i) parameter initialization with a CAE~(see \cref{subsec:CAE_construct}) and trained by optimizing the standard RL1, ii) parameter optimization by iterating between computing an auxiliary target distribution and minimizing the Kullback-Leibler divergence~(KLD)~\cite{KLD} and \emph{cluster assignment hardening loss}~(CAHL) loss in which the cluster assignment is formulated, followed by the centroid updated with backpropagation. 

\begin{figure*}
	\centering
	\includegraphics[width=0.95\textwidth,height=60mm]{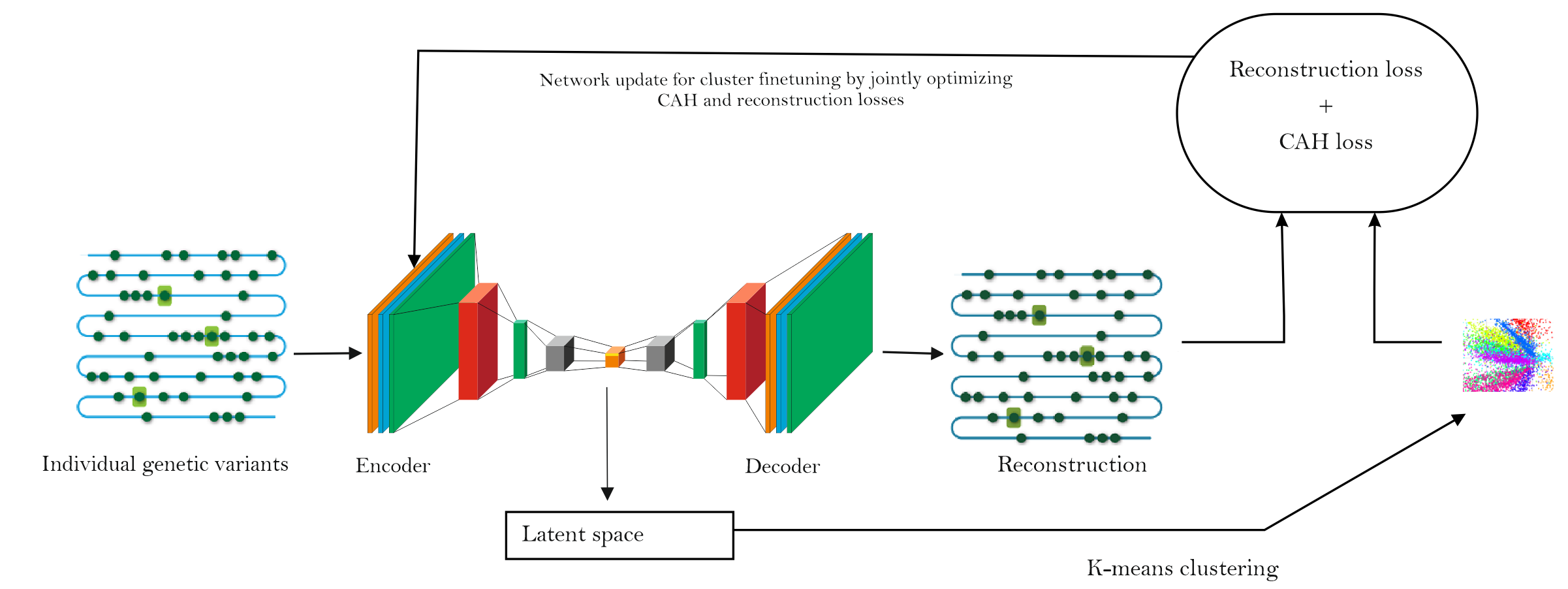}	
    \caption{Improving deep embedding clustering based on CAE and optimizing both reconstruction and CAH losses jointly. Less bright GVs~(right side) means reconstruction errors exist.}	
	\label{cdec_wf}
\end{figure*}

The RL1 guarantees the learned representation to preserve important information~(e.g., spatial relationships between features) so that the original input can be reconstructed~\cite{min2018survey}. Once the RL1 is optimized, latent features~(LF) $Z$ are extracted from the encoder, followed by normalizing them such that $\frac{1}{d}||z_i||^{2}_{2}$ is approximately 1, where $d$ is dimension of the feature space $\lbrace{z_{i}\in Z}\rbrace$. Next, from an initial estimate of the non-linear mapping $f_\theta$ and centroids $\lbrace{\mu_{j}\in Z}\rbrace^{K}_{j=1}$~(as trainable weights $Z$), we improve the clustering by alternating between two steps which we repeat until a convergence criterion is met~\cite{dec2016}: 

\vspace{1mm}
\begin{itemize}
    \item \textbf{Step 1:} soft assignment of $Z$ to the cluster centroids. 
    \item \textbf{Step 2:} updating the mapping $f_\theta$ and refining cluster centroids by learning from initial assignments using an auxiliary target distribution. 
\end{itemize}
\vspace{1mm}

Initializing clustering on LF generates second type of loss called CAHL, which is specific to clustering method and clustering-friendliness of the learned representations~\cite{aljalbout2018clustering}. Similar to literature~\cite{dec2016}, we consider normalized similarities between data points and centroids as the soft cluster assignments in which the Student t-distribution~\cite{maaten2008visualizing} is used as the kernel to measure the similarity between embedded point $z_j$ and centroid $\mu_j$, where $z_i$= $f_\theta$ $(x_i) \in Z$ corresponds to $x_i$ $\in X$ after embedding, $\alpha$ is the degree of freedom, and $q_{ij}$ is the probability of assigning sample $i$ to cluster $j$~\cite{xie2016unsupervised}.

\begin{equation}
    q_{ij}=\frac{(1+||z_{i}-\mu_{j}||^{2}/\alpha)^{-\frac{\alpha+1}{2}}}{\sum_{j'}(1+||z_{i}-\mu_{j'}||^{2}/\alpha)^{-\frac{\alpha+1}{2}}}
    \label{eq:q}
\end{equation}

\noindent However, cross-validation of $\alpha$ in the unsupervised setting is not a viable option. Moreover, learning $\alpha$ is superfluous, similar to literature~\cite{dec2016}, we set $\alpha$ to 1. In step 2, similarity between the distributions is evaluated using KLD w.r.t. by decreasing the distance between soft assignments~($q_{ij}$) and the auxiliary distribution~($p_{ij}$) as follows~\cite{xie2016unsupervised}: 

\begin{equation}
    L_{KLD}=\text{KL}(P||Q)=\sum_{i}\sum_{j}p_{ij}\log\frac{p_{ij}}{q_{ij}},
    \label{eq:KL}
\end{equation}

\noindent where $q_{ij} \in Q$ and $p_{ij} \in P$ are optimized through backpropagation. Minimizing this loss w.r.t. network parameters leads to smaller distances between the data points and their assigned cluster center for a better CQ, where the loss is computed by favoring a situation where points of a cluster are close to the mean of the cluster. Conversely, points that are close to the mean of another cluster will adversely affect the loss. However, since setting $P$ is crucial to increase the CQ, similar to~\cite{dec2016}, the soft assignment $q_{ij}$ is computed by raising auxiliary distribution $p_{ij}$ to the second power and normalizing by frequency per cluster as follows:  

\begin{equation}
    p_{ij}=\frac{q^{2}_{ij}/f_{j}}{\sum_{j'}q^{2}_{ij'}/f_{j'}},
    \label{eq:p}
\end{equation}

\noindent where $f_{j}=\sum_{i}q_{ij}$ are soft cluster frequencies and $P$ forces the assignments to have stricter probabilities between [0--1]. On the other hand, since the constraints enforced by the RL1 can be lost after training the network longer, using only clustering loss may lead to worse clustering results~\cite{aljalbout2018clustering}. To tackle this issue and similar to literature~\cite{min2018survey,DEN,IMSAT}, we performed joint training by setting $\alpha$ such that the network training is affected by both clustering and non-clustering loss functions simultaneously in which combining them with a linear combination of individual loss is priory:

\begin{equation}
    L(\delta) = \sigma L_{KLD}(\delta) + (1 - \sigma)L_{CAE}(\sigma),
    \label{loss_comb}
\end{equation}

\noindent where $L_{KLD}(\delta)$ is the clustering loss, $L_{CAE}(\sigma)$ is the non-clustering loss, and $\sigma \in [0, 1]$ is a constant hyperparameter to specify the weighting between both functions. We optimize $L(\delta)$ using the first-order gradient-based AdaGrad with varying learning rates and different batch size, where gradients of $L$~(w.r.t $Z$) for each data point $z_i$ and cluster centroid $\mu_{j}$ are computed as follows~\cite{dec2016}:

\begin{eqnarray}
    \frac{\partial L}{\partial z_{i}}&=&\frac{\alpha+1}{\alpha}\sum_{j}\left(1+\frac{||z_{i}-\mu_{j}||^{2}}{\alpha}\right)^{-1}\\
    & &\times(p_{ij}-q_{ij})(z_{i}-\mu_{j})\nonumber\\
    \frac{\partial L}{\partial \mu_{j}}&=&-\frac{\alpha+1}{\alpha}\sum_{i}\left(1+\frac{||z_{i}-\mu_{j}||^{2}}{\alpha}\right)^{-1} \\
    & &\times(p_{ij}-q_{ij})(z_{i}-\mu_{j}) \nonumber.
\end{eqnarray}

\noindent where the gradients $\partial L$/$\partial z_{i}$ are used in standard backpropagation to compute network's parameter gradient $\partial L$/$\partial \theta$. This iterative process continues until less than $tol$\% of points change cluster assignment between two consecutive iterations for the cluster assignments~\cite{karim2018recurrent}. 

\subsubsection{Training of CAE classifier}
On the other hand, the CAE classifier has two module: autoencoder and classifier. As discussed in the previous section, after training the CAE, we remove the decoder components by making the first 20 layers trainable false, since the encoder part is already trained. 80\% of the LF is used for the training the CAE classifier and 30\% as the held-out test set. The training LF vector is fed into a flattening layer, followed by a dense, dropout, and Softmax layers~(with output unit of 26 for 1000GP and 7 for SGDP) for the probability distribution over the classes. 

\begin{figure*}[h]
	\vspace{-2mm} 
	\centering
	\includegraphics[width=\textwidth,height=50mm]{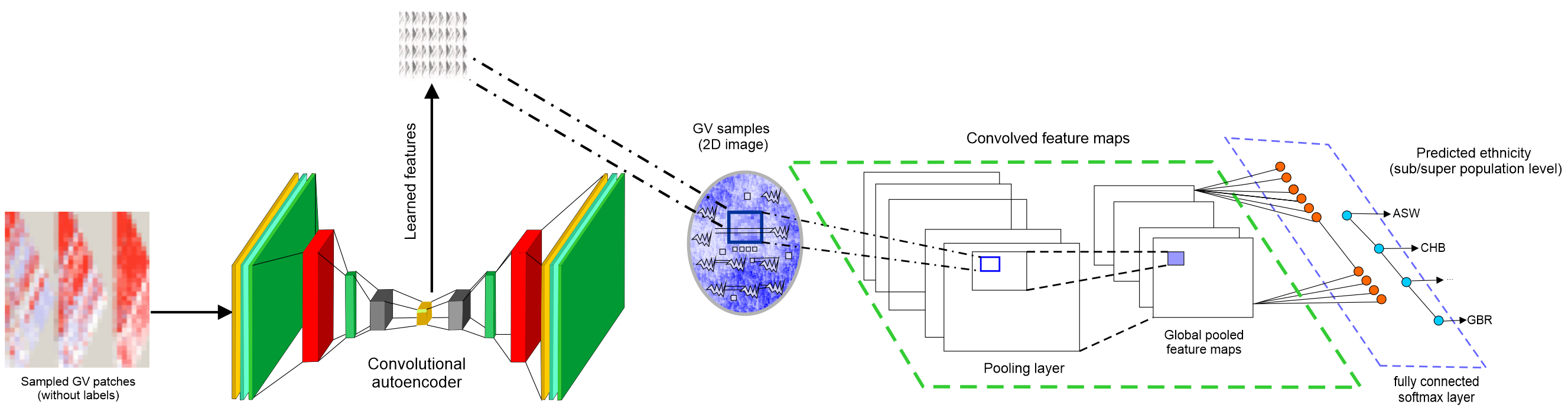}
    \caption{schematic representation of CAE classifier starts from taking input GVs and passing to CAE before getting the latent representation from the encoder to pass to dense, dropout, and softmax layer for the super~(or sub) population prediction}	
	\label{fig:clstm}
	\vspace{-2mm} 
\end{figure*}

Gaussian noise layer was also added followed by each conv, dense, and dropout~(with a high probability value) to improve the model generalization by reducing overfitting. Moreover, to improve the classification results, we applied batch normalization and kept the adaptive rate. Weights are then updated using backpropagation, whereas the AdaGrad optimizer is used to optimize the categorical cross-entropy loss between predicted~(P) vs. true sub-population groups~(T) in a random-search and 5-fold cross validation setting for finding optimal hyperparameters: 

\begin{align} 
    E=\sum_{i,j} T_{i,j} \log P_{i,j}+\left(1-T_{i,j}\right) \log \left(1-P_{i,j}\right)
    \label{eq:cce}
\end{align} 

\noindent When the training is completed, the model is used to score against the test set to measure predicted population groups versus GVs producing in a multiclass setting.


\subsection{Formulating performance metrics}
We used three empirical measures: Elbow~\cite{elbow}, generalizability $G$, and normalized mutual information~(NMI)~\cite{strehl2002cluster}. 
In Elbow, we calculate the cost using within-cluster sum of squares~(WCSS) as a function of the number of clusters, $K$. Since Elbow performs better in a classical clustering setting~\cite{karim2018recurrent}, NMI is used for evaluating clustering results with different cluster numbers~\cite{xie2016unsupervised}, which tells us the reduction in the entropy of class labels, computed as follows: 

\begin{equation}
    NMI(y, c) = \frac{I(y,c)}{\frac{1}{2}[H(y) + H(c)]},
    \label{eq:nmi}
\end{equation}

\noindent where $y$ signifies the ground-truth labels, $c$ is the cluster assignment, $I$ is the mutual information between $y$ and $c$, and $H(.)$ is the entropy. On the other hand, $G$ is the ratio between training and validation loss \cite{xie2016unsupervised}, in which $G$ is small when training loss is lower than the validation loss, an indication of high degree of overfitting. 

\begin{equation}
    G = \frac{L_{train}}{L_{validation}}.
    \label{eq:g}
\end{equation}

\noindent Since a good clustering accuracy also characterized by high intra-cluster similarity and low inter-cluster similarity for the data points. Accordingly, rand index~(RI) is calculated based on the permutation model~(PM) as follows~\cite{karim2020BiB}:

\begin{equation}
    RI = \frac{TP+TN}{TP+FP+FN+TN},
    \label{eq:ri}
\end{equation} 
\vspace{-1mm} 

\noindent where TP, TN, FP, and FN signify true positive, true negative, false positive, and false-negative rates. RI has a value between 0 and 1, where 0 indicates the disagreement between two data clusters on any pair of points, and 1 signifies the perfect agreement~(i.e., the same cluster). In our approach, normalized RI~(ARI), which ranges between -1~(independent labeling) and 1~(perfect match) is used~\cite{rand1971objective}.
\noindent Further, to evaluate the CQ, unsupervised clustering accuracy~(ACC)~\cite{xie2016unsupervised} metric is used, which  takes a cluster assignment from a base clustering algorithm, assigns the ground truths, and computes the best match between them. From the ground ground-truth label $y_i$ and the cluster assignment $c_i$, ACC is computed as follows:

\begin{equation}
    ACC = \operatorname*{max}_{m} \frac{\sum_{i=1}\limits^n 1\Bigl\{y_i=m(c_i)\Bigr\}} {n},
    \label{eq:acc}
\end{equation}

\noindent where $m$ ranges overall possible one-to-one mappings between clusters and labels using Hungarian algorithm~\cite{kuhn1955hungarian}. Further, since ground truths are available, homogeneity and completeness are formulated to desirable objectives for the cluster assignment using conditional entropy analysis~\cite{rosenberg2007v}. While, the former signifies if each cluster contains only members of a single class, the latter tells us if all members of a given class are assigned to the same cluster.

\section{Experiment Results}
\label{section4}
We discuss the evaluation results, both quantitatively and qualitatively. Besides, a comparative analysis with state-of-the-art approaches is provided. 

\subsection{\textbf{Experiment setup}}
All the programs\footnote{\url{https://github.com/rezacsedu/Convolutional-embedded-networks}} were written in Python and experimented on a computer having 32 cores, 256GB of RAM, and Debian 9.9, while the software stack consisted of Keras and scikit-learn with the TensorFlow backend. First, to achieve massive scalability while processing genotype data across all the chromosomes, we used ADAM\footnote{\url{https://github.com/bigdatagenomics/adam}}, while sparkling water\footnote{\url{https://www.h2o.ai/sparkling-water/}} is used to transform data between ADAM and Spark. Network training is then carried out on an Nvidia Titan Xp GPU with CUDA and cuDNN enabled. 

Results based on best hyperparameters produced through random search are reported empirically, where we verified whether the network converges to the optimal number of clusters by setting $K=2$ and increasing it slowly. We also focused on investigating how the network training converged during the cluster assignments and updates by utilizing the Elbow method in which WCSS is calculated. Besides, other metrics such as ARI, NMI, ACC, completeness, and homogeneity. On the other hand, macro-averaged precision, recall, F1-score, and Matthias correlation coefficient~(MCC) are reported in the multiclass setting.

\subsection{\textbf{Ethnicity prediction and inferencing analysis}}
We evaluated performance of the trained models in 3 folds: first, we extracted GVs of selected sub-population groups~(i.e. `ASW', `CHB', `CLM', `FIN', `GBR') from chromosome 22 data, which gives 494,328 alleles allowing 5 class and compare the actual labels to the same number of predicted ethnicity labels. This random sample selection provides a good classification accuracy~(i.e., 92.86\%), as shown in the confusion matrix in~\cref{table:conf}. The full test set is then used to evaluate the model by measuring the prediction performance for sub-populations at the boundaries of the super population, giving precision, recall, F1, and MCC scores of 0.9025, 0.8983, 0.9004, and 0.8245, respectively. 

\begin{table}
\centering
	\caption{confusion matrix of CAE classifier} 
	\label{table:conf}
	\vspace{-3mm} 
	\begin{tabular}{l|l|l|l|l|l|l}
		\hline
		\textbf{Sub population} & \textbf{ASW} & \textbf{CHB} &	\textbf{CLM} & \textbf{FIN} & \textbf{GBR} & \textbf{Support} \\
		\hline
		ASW	& 56 & 0 & 3 & 2 & 0 & 5/61 \\
		CHB	& 2	& 98 & 0 & 3 & 0 & 5/103 \\ 
		CLM	& 1	& 0	& 88 & 4 & 1 & 6/94 \\ 
		FIN	& 1	& 2	& 91 & 3 & 2 & 8/99 \\
		GBR	& 2	& 1	& 1	& 4	& 83 & 8/91 \\
		\hline
		Total & 62 & 103 & 174 & 16 & 85 & 32/448 \\
		\hline
	\end{tabular}
\end{table}

However, since classes are imbalanced, accuracy gives a distorted estimation of the sub-populations. Hence, class-specific classification reports and MCC scores are reported in \cref{table:class_specific}. As shown, precision, recall, and f1 for the majority of sub-populations groups are high. In particular, the CAE classifier classifies JPT, KHV, CEU, TSI, FIN, GBR, IBS, LWK, ASW, ACB, MXL, PUR, CLM, PEL, GIH, PJL, BEB, STU, and ITU samples mostly correctly. However, for the CHB~(Han Chinese in Beijing), CHS~(Southern Han Chinese), and CDX~(Chinese Dai in Xishuangbanna) populations, the CAE classifier made considerable misclassification, giving lower scores, probably because of similar GVs across those samples. Although African populations were mostly classified correctly, in the case of YRI~(Yoruba in Nigeria), GWD~(Gambian in Gambia), MSL~(Mende in Sierra Leone), and ESN~(Esan in Nigeria), a fairly high misclassification error is observed. 

\begin{table}[h]
\centering
    \caption{Class-specific performance of ethnicity prediction}
    \label{table:class_specific} 
    \vspace{-5mm}
    \begin{center}
        \begin{tabular}{l|l|l|l|l}
        \hline
        \textbf{Sub-population} & \textbf{Precision} &  \textbf{Recall}  & \textbf{F1} & \textbf{MCC}\\ \hline
        CHB   & 0.8153 & 0.8015 & 0.8083 & 0.7235  \\\hline
        JPT   & 0.9037 & 0.8976 & 0.90.07 & 0.8256  \\\hline
        CHS   & 0.8233 & 0.8175 & 0.8204 & 0.7421  \\\hline
        CDX   & 0.8025 & 0.7945 & 0.7985 & 0.7043  \\\hline
        KHV   & 0.8643 & 0.8525 & 0.8583 & 0.8046  \\\hline
        CEU   & 0.8583 & 0.8311 & 0.8445 & 0.7967  \\\hline
        TSI   & 0.8826 & 0.8782 & 0.8804 & 0.8024  \\\hline
        FIN   & 0.9235 & 0.9169 & 0.9202 & 0.8369  \\\hline
        GBR   & 0.8943 & 0.9029 & 0.8886 & 0.8162  \\\hline
        IBS   & 0.8224 & 0.8173 & 0.8242 & 0.7121  \\\hline
        YRI   & 0.8381 & 0.8456 & 0.8419 & 0.7235  \\\hline
        LWK   & 0.8967 & 0.9123 & 0.9044 & 0.8531  \\\hline
        GWD   & 0.8194 & 0.8085 & 0.8140 & 0.7891  \\\hline
        MSL   & 0.8368 & 0.8245 & 0.8306 & 0.7951  \\\hline
        ESN   & 0.8785 & 0.8643 & 0.8714 & 0.8097  \\\hline
        ASW   & 0.8954 & 0.8832 & 0.8893 & 0.8430  \\\hline
        ACB   & 0.8753 & 0.8671 & 0.8712 & 0.8421  \\\hline
        MXL   & 0.8825 & 0.8733 & 0.8779 & 0.8262  \\\hline
        PUR   & 0.8913 & 0.8719 & 0.8815 & 0.8076  \\\hline
        CLM   & 0.8537 & 0.8611 & 0.8574 & 0.8735  \\\hline
        PEL   & 0.9629 & 0.9567 & 0.9598 & 0.8525  \\\hline
        GIH   & 0.8736 & 0.8722 & 0.8729 & 0.8134  \\\hline
        PJL   & 0.8952 & 0.8845 & 0.8898 & 0.8236  \\\hline
        BEB   & 0.9255 & 0.9123 & 0.9188 & 0.8374  \\\hline
        STU   & 0.8795 & 0.8857 & 0.8826 & 0.8212  \\\hline
        ITU   & 0.8697 & 0.8567 & 0.8632 & 0.8513  \\\hline
        \textbf{Average} &   \textbf{0.9025}    &  \textbf{0.8983} &    \textbf{0.9004}  & \textbf{0.8245}\\ \hline 
    \end{tabular}
    \end{center}
\end{table}

\begin{figure}[h]
	\centering
	\includegraphics[width=\linewidth]{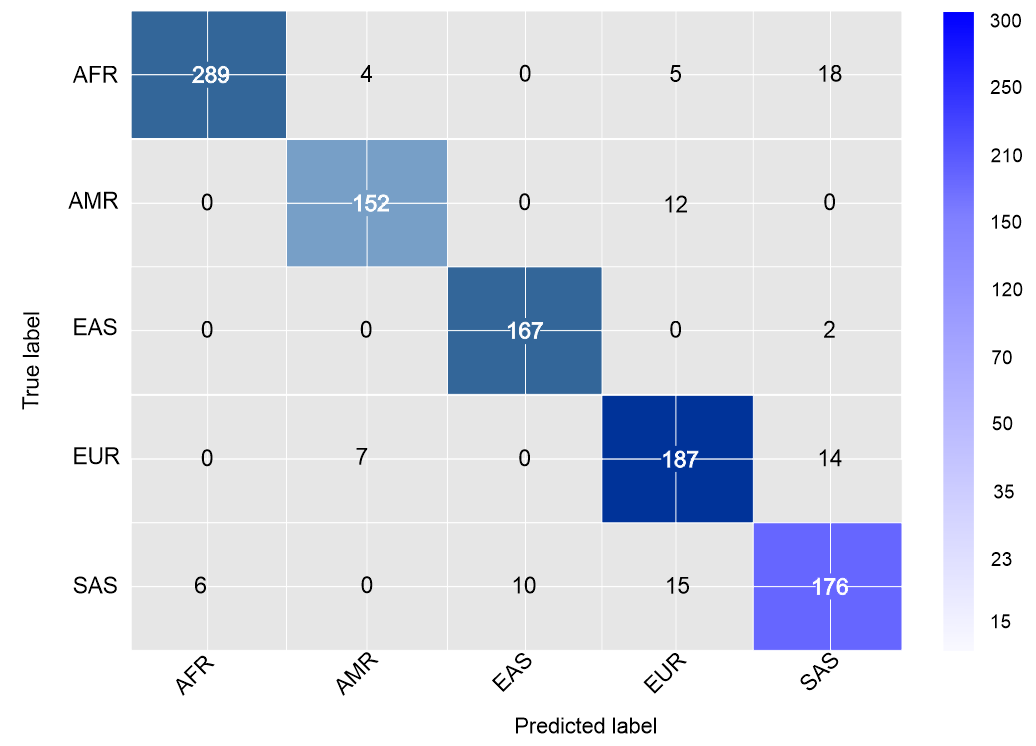}
	\caption{Confusion matrix where predictions are for sub populations at the boundaries of the super population} 
	\label{fig:conf}
\end{figure}

\begin{figure*}
	\centering
	\begin{subfigure}{.48\textwidth}
		\centering
		\includegraphics[width=\linewidth,height=60mm]{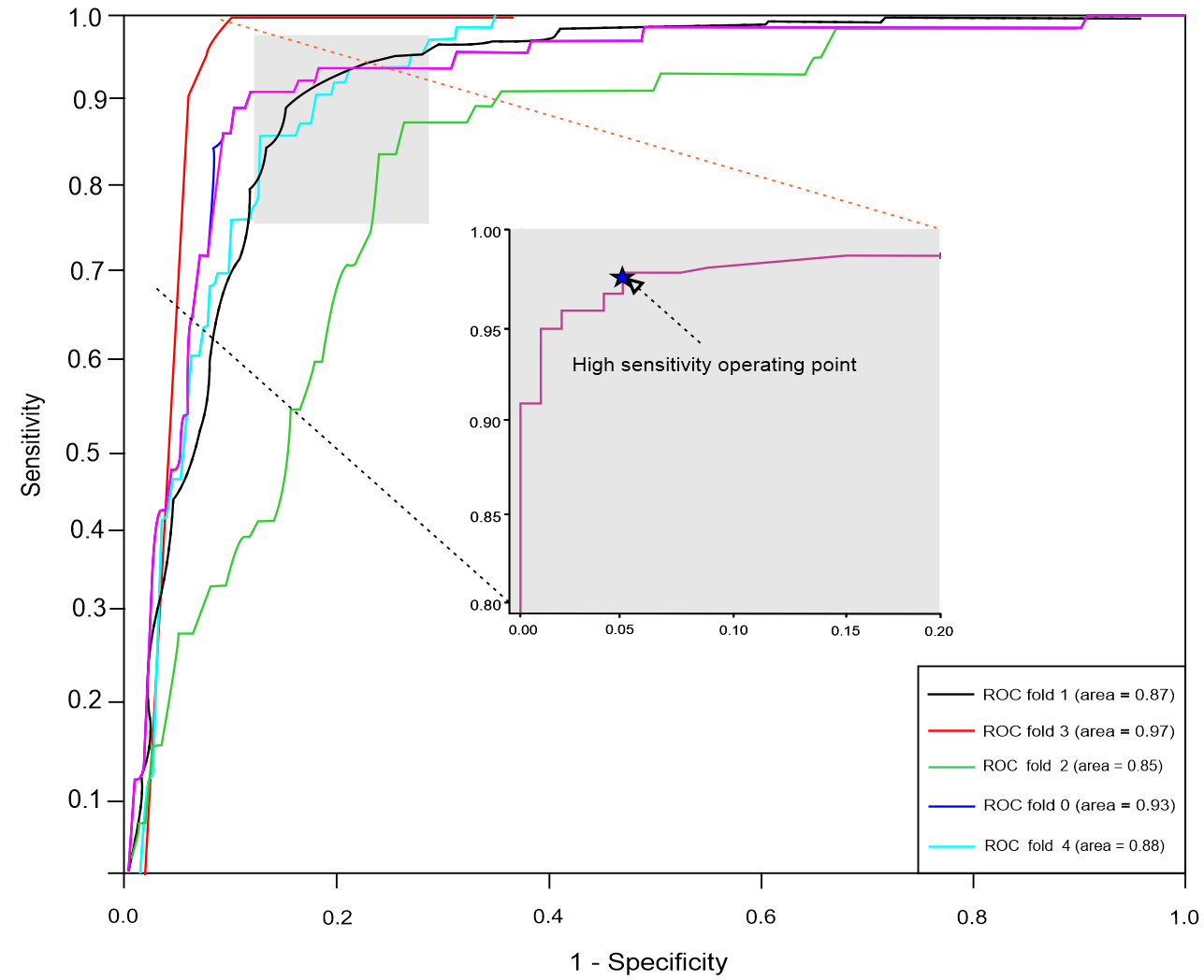}
		\caption{1000GP}
        \label{fig:roc_1000}
	\end{subfigure}%
	\begin{subfigure}{.48\textwidth}
		\centering
		\includegraphics[width=\linewidth,height=60mm]{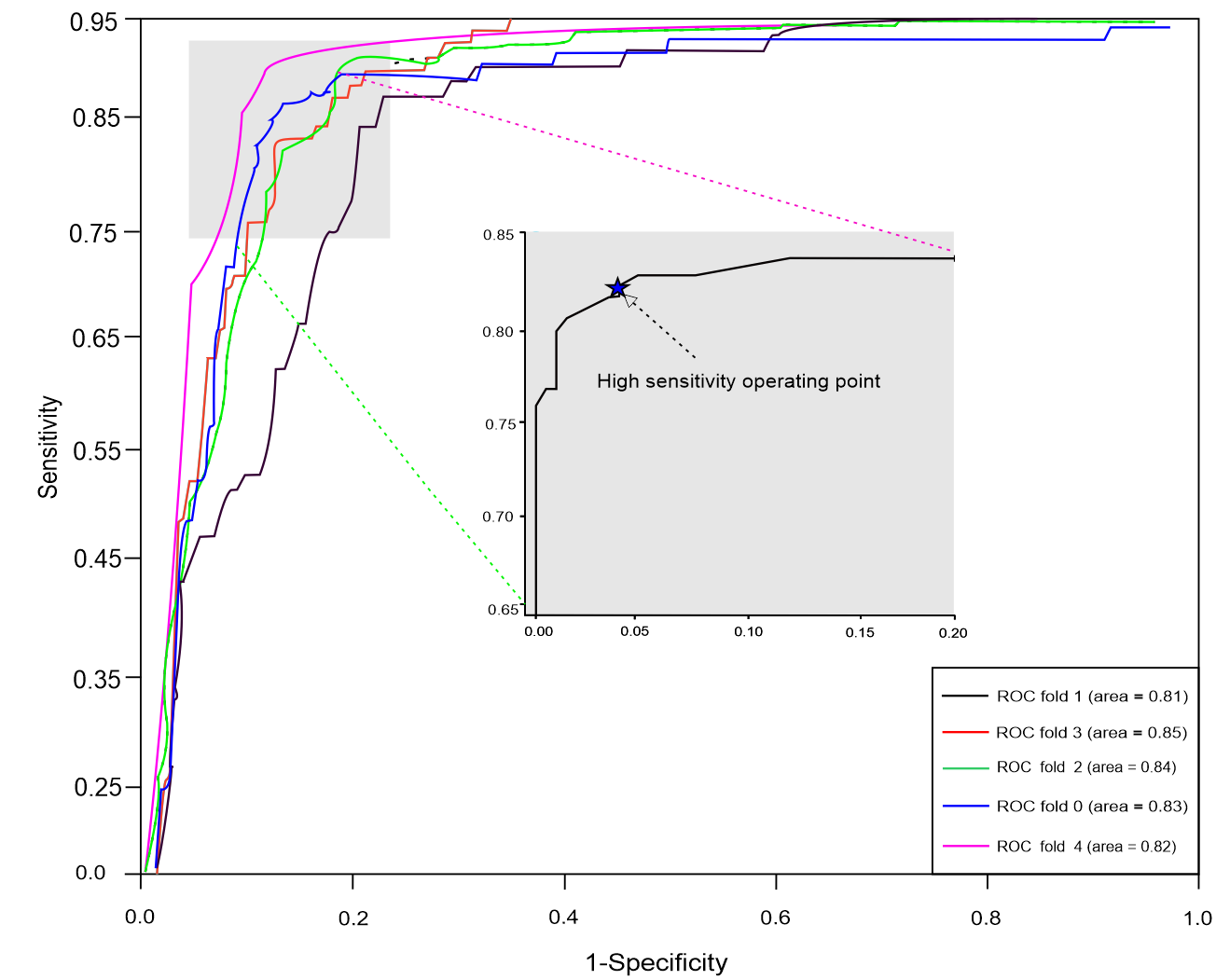}
		\caption{SGDP}
        \label{fig:roc_simon}
	\end{subfigure}
	\caption{ROC curves for the CAE classifier on 100GP and SGDP datasets} 
	\label{roc:both_dataset}
\end{figure*}

Even though, CAE classifier confused between AFR, SAS, AMR, and EUR samples~(as shown in \cref{fig:conf}), this is still considerably low compared to literature~\cite{karimdeep}. The reason for the improvement is that all the variants with high LD and incomplete variants were removed in our preprocessing step, which has contributed towards non-corrupted latent features. The reason is simple with that minor factor, and we removed some impurities giving the network more quality features, which eventually helped in separating data points. On the other hand, class-specific MCC scores of the CAE classifier suggests that predictions were strongly correlated with the ground truth, yielding a Pearson product-moment correlation coefficient higher than 0.70 for the majority sub-population groups. Besides, the ROC curves in \cref{roc:both_dataset} show consistent AUC scores across folds for both datasets, which signifies that the predictions by the CAE classifier in both cases are much better than random guessing. 

Finally, we trained the CAE classifier on the SGDP dataset for predicting unknown ethinicity in which similar data processing steps were followed. We experienced moderately lower predictive accuracy: out of 44 Africans, only 35 samples were predicted correctly, and the rest of the samples were predicted as South Asian. However, the CAE model shows high confidence at predicting Native Americans samples in which 20 samples out 22 were correctly predicted, and only 2 samples were predicted as West Eurasians; in the case of Central Asians or Siberians, 20 samples were correctly predicted, and 7 samples were predicted as West Eurasians, while out of 47 East Asians samples, 39 were correctly predicted, 6 samples were predicted as Africans, and 2 samples were predicted as South Asians. 

In the case of Oceanians samples, 22 samples were correctly predicted out of 25, and only 3 samples were predicted as Native Americans. Besides, out of 39 South Asians samples, 29 were correctly predicted, and 10 samples were predicted as Africans, showing considerably high misclassification rates. Out of 75 West Eurasians samples, only 65 were correctly predicted, 8 were predicted as Central Asians or Siberians, and 2 were predicted as Native Americans.
Overall, 211 samples out of 279 were correctly classified giving an F1-score of 0.83 in which CAE classifier was more confused between South Asian and Africans, Central Asians or Siberians and West Eurasians, and between East Asian and Africans, which is probably because GVs from these two groups are mostly mixed and share common alleles. A depth genomic analysis is further required to explain these. 

\subsection{\textbf{Population scale cluster analysis}}
\begin{figure*}
  \begin{subfigure}[b]{0.5\textwidth}
   \centering
    \includegraphics[width=0.90\textwidth,height=60mm]{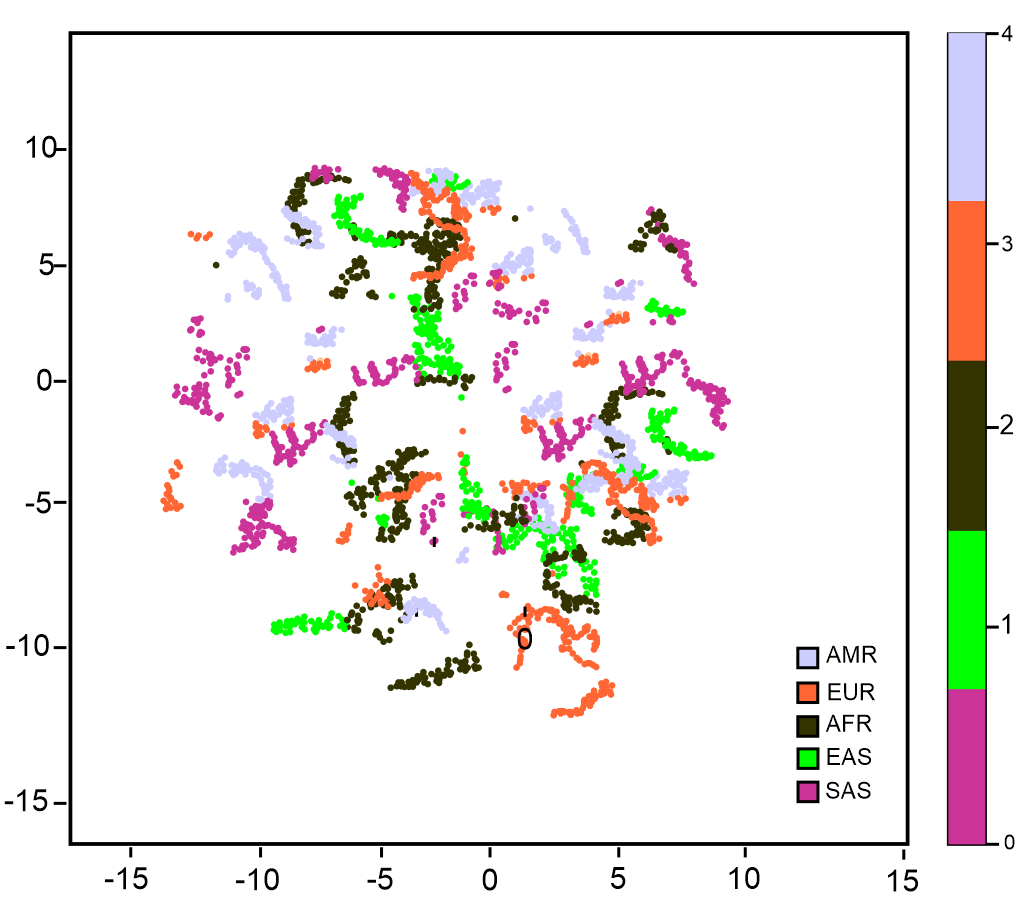}
    \caption{Raw genetic variants}
        \label{fig:raw}
  \end{subfigure}
  \begin{subfigure}[b]{0.5\textwidth}
   \centering
    \includegraphics[width=0.90\textwidth,height=60mm]{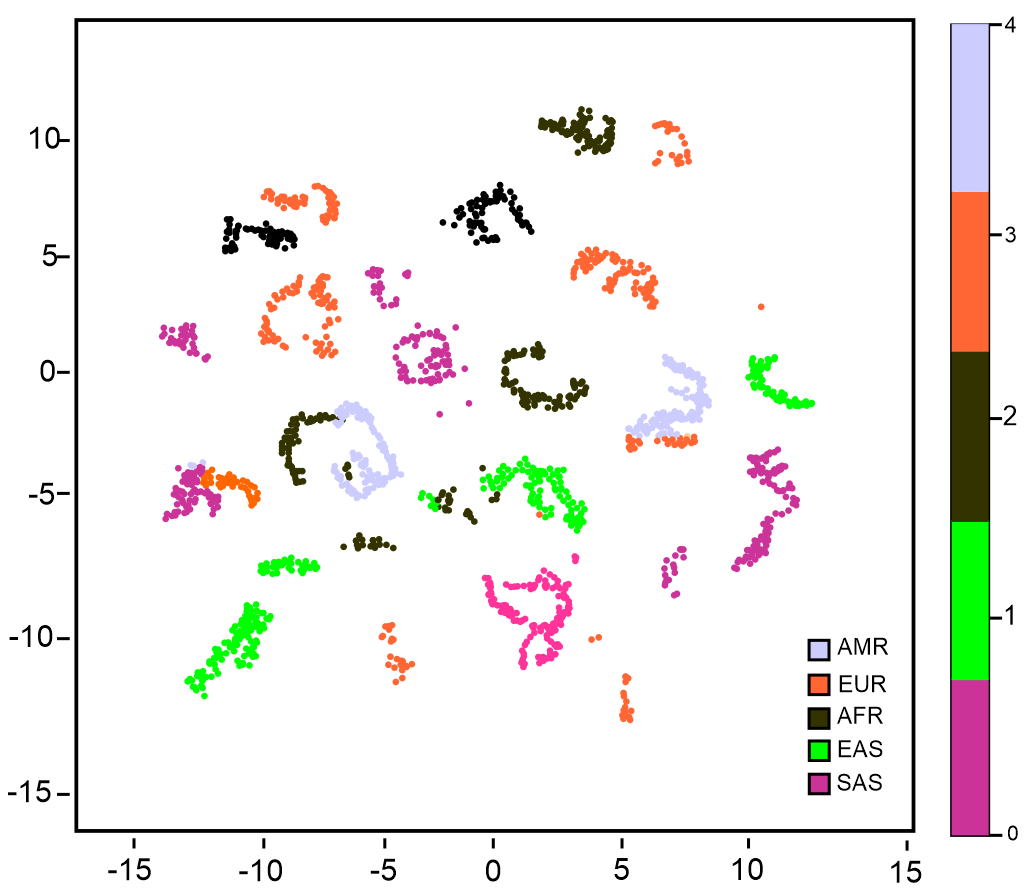}
    \caption{CAE latent FMs}
        \label{fig:cae_lf}
  \end{subfigure}
  \begin{subfigure}[b]{0.5\textwidth}
   \centering
    \includegraphics[width=0.90\textwidth,height=60mm]{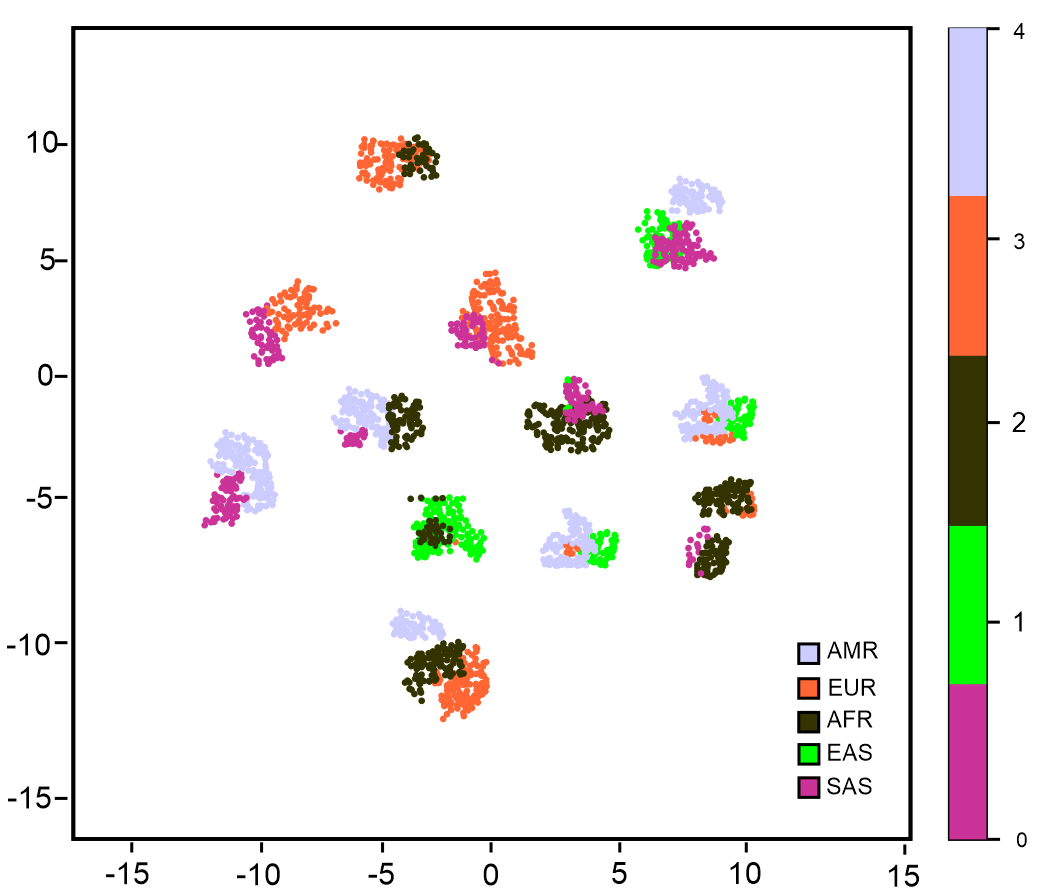}
    \caption{Clustering raw genetic variants with agglomerative clustering}
        \label{fig:ac}
  \end{subfigure}
  \begin{subfigure}[b]{0.5\textwidth}
   \centering
    \includegraphics[width=0.90\textwidth,height=60mm]{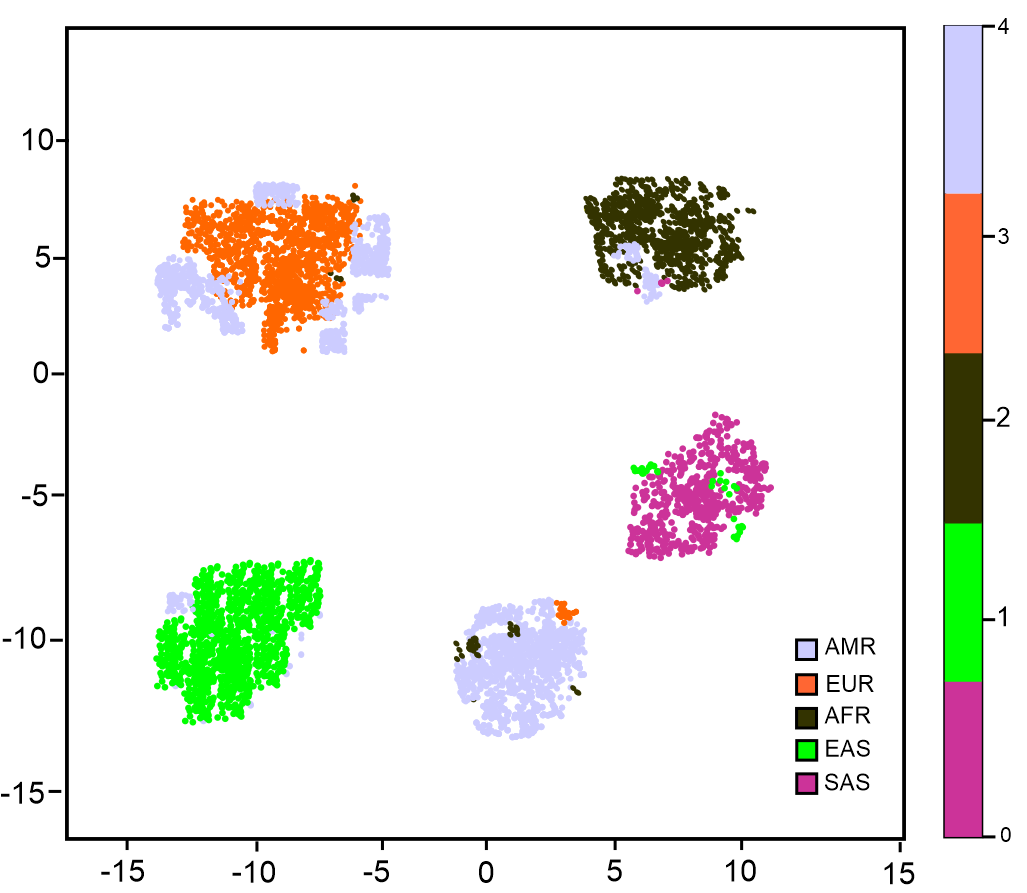}
    \caption{Agglomerative clustering based on CAE latent FMs}
    \label{fig:cae_ac}
  \end{subfigure}
  \caption{t-SNE plots of different stages in clustering breast microscopy images} 
  \label{fig:tnse_image}
\end{figure*}

Clustering results are reported in \cref{table:result} with different metrics in which the AC algorithm performed the best clustering with optimal hyperparameters on CAE-based LF~(highlighted in cyan): we observed an ARI, NMI, and ACC of 0.915, 0.927, and 0.896, respectively in which each cluster contains only members of a single class in 86.7\% of the cases~(homogeneity) and in 85.3\% of the cases all members of a given class are assigned to the same cluster~(completeness). The reason is that CAE learned LF that are more quality ones than raw GVs data, as shown in \cref{fig:tnse_image}, which eventually tends to better separability of data points.

\begin{table}[!htbp]
	\centering
	\caption{clustering with autoencoders + base clustering algorithms~(Hom*==Homogeneity, Com*==Completeness)}
	\label{table:result}
	    \vspace{-2mm}
	\begin{tabular}{l|l|l|l|l|l}
		\hline
        \textbf{Clustering algorithm} & \textbf{ARI} & \textbf{NMI} & \textbf{ACC} & \textbf{Hom*} & \textbf{Com*}\\ \hline
	    K-means & 0.853 & 0.863 & 0.838 & 0.827 & 0.814\\
		AC & {\color{cyan}\textbf{0.915}} & {\color{cyan}\textbf{0.927}} & {\color{cyan}\textbf{0.896}} & {\color{cyan}\textbf{0.867}} & {\color{cyan}\textbf{0.853}} \\
		GMM & 0.814 & 0.821 & 0.792 & 0.775 & 0.776\\
		DBSCAN & 0.885 & 0.891 & 0.872 & 0.846 & 0.836\\
		OPTICS & 0.831 & 0.843 & 0.815 & 0.813 & 0.805\\
		\hline
	\end{tabular}
\end{table}

Besides, DBSCAN and K-means~(on CAE-based on LF) also performed moderately well compared GMM and OPTICS algorithms in which DBSCAN turns out to be the second-best clustering algorithm. Contrarily, both OPTICS and GMM did not perform well, making the GMM the worst performer. On one hand, OPTICS is inherently better for sequences hence couldn't provide better separability of the data points. On the other hand, GMM doesn't have any uncertainty measure or probability w.r.t how much a data point is associated with a specific cluster. 

Predicted clusters by CEC can be observed in \cref{fig:cae_ac}, where GVs of African~(AFR), East Asian~(EAS), and American~(AMR) super-population groups are highly separated and grouped into distinct clusters. On the other hand, although European individuals can be mostly separated into persons of Finnish and non-Finnish ancestry~\cite{padhukasahasram2014inferring}, overall EUR is more mixed with AMR and consists predominantly of individuals, so didn't cluster well. This is probably the effect of several mixtures of GVs, e.g., migrational backgrounds and a potential reason for such low clustering accuracy.  However, in cases where individuals were not clearly clustered due to diverse migrational backgrounds, the ancestry itself influences the treatment hence accurate population association may not be known for the patients even from the human leukocyte antigen~(HLA) allele genotyping from SNP information~\cite{variantspark}. Albeit, literature~\cite{variantspark} observed an increase in cluster accuracy by removing individuals with a mixed background and operating at super-population level, and it was not the cases in the sub-population level sine a recent research~\cite{27} has shown that the apparent separation between East Asian and other samples reflects a deficiency of Middle Eastern and Central Asian samples in the dataset. 

Further, we investigate how well CEC converges to the optimal number of population groups: we started the clustering by setting K=2~(where applicable) and increased up to 35 and observe the clustering performance. We plotted NMI and G vs. several clusters and found a sharp drop of generalizability for K = 26 and 27, which means that 26 is the optimal number of clusters. To support this argument, the graph also shows that for 26, we observed the highest NMI of about 0.92. Subsequently, we utilize the Elbow method in which we calculated WCSS as a function of the number of clusters~(i.e., K), for which we observed a drastic fall of WCSS when several clusters were around 25 and 26 for 1000GP and 7 and 8 for the SGDP. 

To compare with VariationSpark, we further analyze the clustering of five super-population groups~(i.e., `EUR', `AMR', `AFR', `EAS', `SAS') for each to the label assigned by CEC. This experiment results in an ARI of 0.87, an ACC of 0.86, and an NMI of 0.88, showing higher confidence, at least in terms of ARI~(albeit, this is still low compared to our overall ARI). 
Finally, we investigate which super-population group contains what percentage of human GVs in which CEC reveals an interesting statistics showing majority of the genetic variants were clustered into EUR~(28.32\%) and lowest into AMR~(12.68\%), while the distributions of samples from EAS, AFR, and SAS super-population groups were 22.25\%, 18.65\%, and 18.10\%, respectively. 

\subsection{Qualitative study of the learned representations}
Since GVs data are high-dimensional, learning the association between each feature was fairly considered. Inspired by literature~\cite{rhee2017hybrid,KarimIEEEAccess2019,KarimNCCA2019} and to qualitatively study whether the learned representation can express the biological characteristics of the individuals, t-SNE of the CAE encoder's output, i.e., latent FM and raw GVs are plotted in \cref{fig:tnse_image}. From t-SNE plots, we can observe moderately high distinctive patterns between 4 super-population groups. However, not all these patterns clearly visible in the t-SNE plot of raw GVs, which signifies how CAE learned latent genetic properties better from the GVs profiles. 

\begin{figure*}
\centering
	\includegraphics[width=0.8\linewidth]{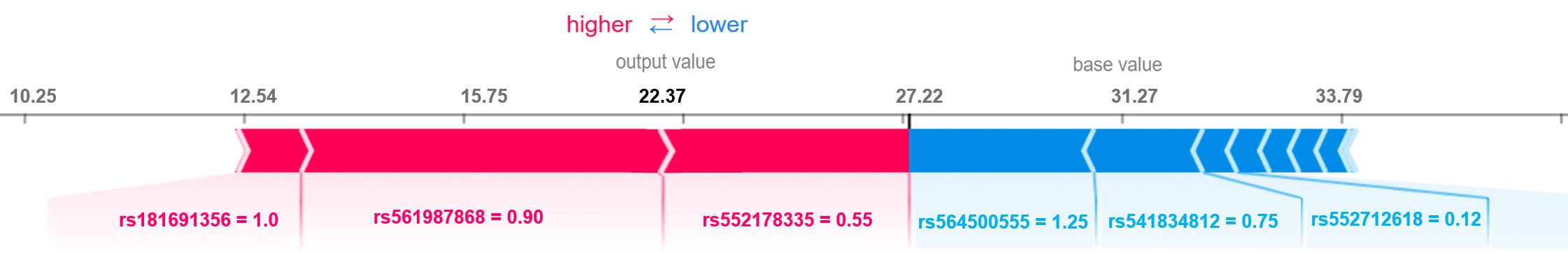}
	\caption{GVs feature contribution for the first prediction: features pushed the prediction higher and lower in red and blue} 
	\label{fig:shap}
\end{figure*}

\subsection{Shapley value-based explanations}
Since CAE learns the representations are not easily interpretable, interpretable and disentangled representations are essential to provide insight into what features did the representations capture and what attributes of the samples are the clusters based on. To provide interpretations of the predictions, we identify significant GV biomarkers using GBT in which SHAP generates explanations. Shapley values are used to calculate the importance of a feature by comparing what a model predicts with and without a feature from all possible combinations of $n$ features in the dataset $S$, where the prediction $p$ of the model for a given feature $i \in S$ is generated w.r.t $i$, for which the Shapely value $\phi$ is calculated as follows~\cite{NIPS2017_7062}: 

\vspace{-3mm}
\begin{align}
    \phi_{i}(p)=\sum_{S \subseteq N / i} \frac{|S| !(n-|S|-1) !}{n !}(p(S \cup i)-p(S))
    \label{eq:shap}
\end{align}
\vspace{-3mm}

However, since the order in which a model sees features can affect the predictions, this computation is repeated in all possible orders to compare the features fairly~\cite{karim2019onconetexplainer}. The feature that does not modify the predicted value is expected to produce a Shapley value of 0. However, if two features contribute equally to the prediction, Shapley values should be the same~\cite{NIPS2017_7062}. The base value indicating the directions of the first prediction made by the GBT model is \cref{fig:shap} in which how much each feature is pushing model's output from the base value\footnote{The average model output over the training dataset passed} to the predicted output is shown. Features pushing the prediction higher are shown in red, those pushing to lower are in blue. 

\begin{figure}
\centering
	\includegraphics[width=0.7\linewidth]{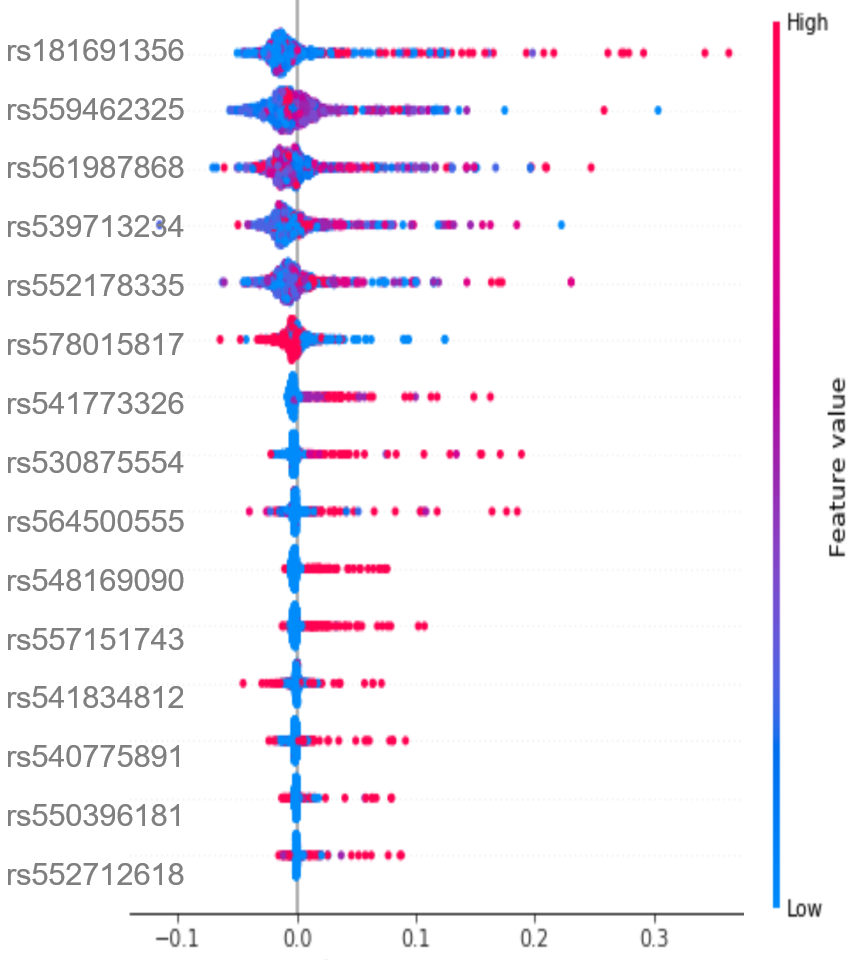}
	\caption{Clinical features ordered by ascending importance on the y-axis~(dots represent SHAP values of specific features)} 
	\label{fig:shap_FI}
\end{figure}

\begin{table}
\centering
	\caption{annotations of top-15 variants}  
	\label{tab:gvs_ann}
	\vspace{-3mm} 
	\begin{tabular}{l|l|l|l}    
    \hline
		\textbf{Variant ID} & \textbf{Variant type} & \textbf{Alleles} &  \textbf{Allele frequency} \\   
        		\hline
		rs181691356 & SNP & C$>$ A,T  & A = 0.0008/4  \\
		rs559462325 & SNP & G$>$ A    & A = 0.0002/1  \\
		rs561987868 & SNP & G$>$ A,T  & T = 0.0002/1  \\
		rs539713234 & SNP & G$>$ C    & C = 0.0046/23  \\
		rs552178335 & SNP & G$>$ A    & A = 0.0002/1   \\
		rs578015817 & SNP & G$>$ A    & A = 0.0002/1  \\
		rs541773326 & SNP & C$>$ G,T  & T = 0.0002/1  \\
		rs530875554 & SNP & C$>$ A,T  & A = 0.001/5  \\
		rs564500555 & SNP & A$>$ C,G  & G = 0.0074/37  \\
		rs548169090 & SNP & G$>$ A    & A = 0.0002/1  \\
		rs557151743 & SNP & T$>$ C    & C = 0.0004/2  \\
		rs541834812 & SNP & G$>$ A    & A = 0.0002/1  \\
		rs540775891 & SNP & C$>$ A,T  & C = 0.0002/1  \\
		rs550396181 & SNP & T$>$ C    & C = 0.0004/2   \\
		rs552712618 & SNP & A$>$ G,T  & T = 0.0006/3  \\
		\hline
	\end{tabular}
\end{table}

In \cref{fig:shap_FI}, top-15 common biomarkers are sorted by the sum of SHAP value magnitudes over all the samples and are ordered according to their importance, where the color represents the value of the feature from low to high, i.e., red represents high feature values, and blue represents low feature values. Overlapping points are jittered in the y-axis signifying the distribution of the Shapley values per feature, i.e., delivery of the impact of each feature on the model output. However, since all effects describe the overall behavior of the model and are not necessarily causal, we provide additional annotations for these biomarkers in \cref{tab:gvs_ann}. As seen, the majority of the variants are multi-allele SNPs. 

\subsection{Discussion and comparative analysis}
Based on optimal K and other hyperparameters, CEC completes clustering in 22 hours with an ARI of 0.915, an NMI of 0.927 and ACC of 0.896 as shown in \cref{tab:compare}, while VariationSpark requires 30h to finish the overall computation, leveraging an ARI of 0.82 only\footnote{VariationSpark and ADMIXTURE did not report NMI, ACC, homogeneity, and completeness}. 

\begin{table}[htp!]
\centering
	\caption{clustering result comparison between different approaches~(Hom*==Homogeneity, Com*==Completeness)}  
	\label{tab:compare}
	\vspace{-3mm} 
	\begin{tabular}{l|l|l|l|l|l|l}    
    \hline
		\textbf{Approach} & \textbf{ARI} & \textbf{NMI} & \textbf{ACC} & \textbf{Hom*} & \textbf{Com*} & \textbf{Time} \\   
        		\hline
		CEC & 0.915 & 0.927 & 0.896 & 0.867 & 0.853 & 22h \\
		        \hline
        VariantSpark & 0.82 & - & - & - & - & 30h \\
        		\hline 
      ADMIXTURE	& 0.25 & - & - & - & - & 35h \\
		\hline
	\end{tabular}
\end{table}

ADMIXTURE performs clustering based on the maximum likelihood estimation~(MLE) of individual ancestries and multi-locus SNP genotypes. Overall processing time is considerably high~(takes 35h), giving an ARI of only 0.25~\cite{21}. Following are some potential reasons behind low clustering accuracy of VariationSpark and ADMIXTURE:

\vspace{1mm}
\begin{itemize}
    \item Being a K-means-based approach, VariationSpark has several limitations. K-means is based on the assumption that each cluster is equal-sized, where clusters to have hyper-sphere shapes~\cite{karim2020BiB}, which is not the case for both 1000GP and SGDP datasets. Further, since K-means is sensitive to noise and outliers thus was probably trapped in a local optimum during the clustering operations. Nevertheless, we observed that the clustering results were slightly different for a different initial value of K since it's the only hyperparameter. 

    \item Being an MLE-based approach, ADMIXTURE is limited to accurately estimate population mean and standard deviation~\cite{26} in case of multi-locus SNP genotypes. 
\end{itemize}
\vspace{1mm}

Our study also investigated what percentage of the cases for each cluster contains members of a single class and what percentage of the cases for all members of a given class are assigned to the same cluster, giving 86.7\% of homogeneity and 85.3\% of the completeness of the clustering. 
Further, the in-memory caching mechanism of ADAM and Spark while processing VCF files made our processing pipeline 32\% and 90\% faster compared to VariantSpark and ADMIXTURE, respectively. Nevertheless, VariationSpark and ADMIXTURE are both black-box methods, whereas we explain the prediction made by models showing class and cluster discriminating features.

\section{Conclusion and Outlook}
\label{section6}
In this paper, we implemented two powerful architecture called CEC and CAE classifiers for the clustering population and predicting genomic ancestry based on GVs of about 3,000 individuals from the 1000GP and SGDP. Our Spark and ADAM based data processing is particularly suitable for handling large-scale genomic data. Experiment results with a focus on accuracy and scalability show that our approach outperforms state-of-the-art approaches such as ADMIXTURE and VariantSpark. 
Our approach can perform clustering on VCF files from 2504 individuals consist of 84 million GVs in just 22h, allowing faster clustering for well-characterized cohorts, where 20\% of the genome is sufficient for the training. CEC can cluster the whole population by jointly optimized feature space with an ACC of 89\%, which can be viewed as an unsupervised extension of semi-supervised self-training. Similar to~\cite{xie2016unsupervised}, CEC has linear complexity concerning several data points, which allowed us to scale to large datasets, whereas the CAE classifier can predict the ethnicity of unknown samples with an F1-score of 93\%, which is consistent with all the genotypic dataset from 23 chromosomes, giving high-level of confidence. 

We have seen that explaining the predictions with plots and charts are useful for exploration and discovery but interpreting them for the first time may be tricky, e.g., suppose, the CAE classifier predicts~(or groups) a selected sample is of FIN population~(or into EUR super-population group) in which model’s average response to the dataset is 0.6 and the model predicts that the selected sample's bio-ancestry with probability 0.75 by showing all the variables that have contributed to that prediction but still interpreting these for the first time may be tricky and need more human-interpretable decision rules in natural language. 
In the future, we intend to extend this work by providing: i) a more detailed analysis of intra-super-population groups and discuss the homogeneity and heterogeneity among different groups, ii) considering other datasets and factors like predicting population groups within larger geographic continents, iii) exploring if we can make share representations of the features out of both 1000 genomes and PGP datasets and cluster them simultaneously using CEC, iv) by generating decision rules to provide more human-interpretable explanations using neuro-symbolic reasoning. 

\bibliographystyle{IEEEtran}
\bibliography{Main}

\end{document}